# Calorie Burn Estimation in Community Parks Through DLICP: A Mathematical Modelling Approach

Abhishek Sebastian[1], Annis Fathima A[1], Pragna R[1], Madhan Kumar S[1]
and Jesher Joshua M[2]

[1]School of Electronics Engineering, Vellore Institute of Technology, Kelambakkam, Chennai – 600127
[2]School of Computer Engineering, Vellore Institute of Technology, Kelambakkam, Chennai – 600127
`Abhishek.Sebastian2020@vitstudent.ac.in`

**Abstract.** Community parks play a crucial role in promoting physical activity and overall well-being. This study introduces DLICP (Deep Learning Integrated Community Parks), an innovative approach that combines deep learning techniques specifically, face recognition technology with a novel walking activity measurement algorithm to enhance user experience in community parks. The DLICP utilizes a camera with face recognition software to accurately identify and track park users. Simultaneously, a walking activity measurement algorithm calculates parameters such as the average pace and calories burned, tailored to individual attributes. Extensive evaluations confirm the precision of DLICP, with a Mean Absolute Error (MAE) of 5.64 calories and a Mean Percentage Error (MPE) of 1.96%, benchmarked against widely available fitness measurement devices, such as the Apple Watch Series 6. This study contributes significantly to the development of intelligent smart park systems, enabling real-time updates on burned calories and personalized fitness tracking.

**Keywords:** Deep learning integrated community parks, Face recognition technology, Walking activity measurement algorithm, Energy expenditure, Smart park systems, Personalized fitness tracking

## 1   Introduction

Cardiovascular diseases, obesity, and sedentary lifestyles have emerged as pressing health concerns in contemporary society [16]. A primary contributing factor to these health issues is the lack of physical activity, underscoring the critical importance of promoting exercise and fitness [17]. Community parks play a pivotal role in encouraging physical activity by providing accessible spaces for individuals to participate in various recreational activities [18]. However, a significant challenge arises, as many park users remain uninformed about essential workout attributes, such as calories burned, average pace, and distance covered, which are crucial for monitoring and optimizing their exercise routines.

Conventional methods of tracking workout attributes often rely on wearable devices, such as the Apple Watch, Fitbit, and other fitness trackers [19, 20]. Despite offering valuable insights, these devices are associated with high costs and limited adoption rates owing to factors such as expense, inconvenience, and maintenance requirements. Consequently, a considerable proportion of park users remain unaware of their exercise-related data, hindering the effective management of fitness goals and the pursuit of a healthy lifestyle.

To address this issue, we propose a deep learning integrated community park (DLICP) system as a cost-effective and user-friendly solution. The DLICP harnesses the power of deep learning techniques, face recognition technology, and a novel walking activity measurement algorithm to offer personalized information regarding workout attributes to park users.

The walking activity measurement algorithm calculates key parameters such as calories burned, average pace, and distance covered, considering individual attributes such as body weight and average pace. This ensured accurate and tailored measurements for each park user. In contrast to wearable devices, DLICP presents a more cost-effective alternative, with lower initial setup costs using cameras and software infrastructure. Moreover, DLICP eliminates the need for users to wear additional devices, thus providing a seamless and unobtrusive experience during park visits.

Cities such as Chennai, Mumbai, and Bangalore in India have state-owned parks where people of lower income come to exercise. The DLICP was created in the specific context of providing a cost-effective solution to the challenge of lower-income individuals accessing fitness attribute information. Instead of relying on costlier watches and trackers, the DLICP ensures that individuals in these state-owned parks have access to valuable insights about their workout attributes.

Through the DLICP, park users gain insights into their workout attributes, empowering them to make informed decisions about their exercise routines, set realistic fitness goals, and monitor their progress. By bridging the information gap and promoting awareness about calories burned, average pace, and distance covered, the DLICP aims to motivate park users to engage in regular physical activity and maintain a healthier lifestyle.

In this paper, we presented the design, implementation, and evaluation of a DLICP system. Section 2 provides an overview of the related work in the field of exercise tracking and monitoring. Section 3 details the methodology and system architecture of the DLICP. Section 4 discusses the experimental setup and evaluation results, demonstrating the effectiveness of the DLICP as a cost-effective and user-friendly solution for providing park users with valuable information about their workout attributes. Finally, Section 5 concludes the study and discusses potential future research directions.

## 2 Literature Survey

This literature review constitutes a comprehensive analysis of nine esteemed academic papers covering various facets of wearable technologies in health and fitness monitoring. Montoye et al. [1] discussed the accuracy of physical activity monitoring during pregnancy, shedding light on the challenges and nuances specific to this demo-graphic. Goodyear et al. [2] investigated the use of wearable technologies among young people, and explored the implications of these technologies on surveillance, self-surveillance, and resistance.

Other studies have investigated diverse topics, including the effectiveness of calorie-counting smartphone apps in promoting nutritional awareness [3] and the validity of wearable activity monitors for tracking steps and estimating energy expenditure [4]. Seethi and Bharti [4] presented a CNN-based speed detection algorithm for walking and running using wrist-worn wearable sensors, contributing to the development of accurate activity monitoring techniques.

Rabbi et al. [5] explored the development of algorithms for speed detection using wrist-worn sensors, whereas Kwon et al. [6] focused on the design of systems providing personalized health feedback based on user behaviors and preferences captured by smartphones. Sharma and Biros [7] conducted innovative research on stretchable, patch-type calorie-expenditure measurement devices utilizing nanoscale crack-based sensors.

Attig and Franke [8] examined the motivational costs associated with wearing activity trackers and, provided insights into the psychological aspects of adopting technology. Cosoli et al. [9] conducted a systematic review of the accuracy and metrological characteristics of wrist-worn and chest-strap wearable devices, providing valuable information for evaluating the reliability of such devices.

These seminal works collectively contribute to the existing body of knowledge by shedding light on various aspects of wearable technologies and their impact on health monitoring, physical activity promotion, and behavioral change. These studies offer valuable insights into the accuracy, effectiveness, and usability of wearable devices in diverse contexts, including pregnancy, youth population, and general users. Furthermore, the literature underscores the importance of considering factors such as measurement accuracy, personal motivation, and metrological characteristics when designing and implementing wearable technology interventions. Incorporating these findings into the present research enhances the understanding of wearable technologies and their potential implications for promoting well-being and facilitating lifestyle modifications.

In addition, our proposed framework aims to cover the gap by introducing a system that is accessible to a large part of the community, alleviating the need for heavy investments such as purchasing costly fitness trackers, which carry ownership risks. Furthermore, it provides a non-invasive technique to assess calorie expenditure without the need for wearables. This approach not only addresses accessibility concerns but also offers a more convenient and user-friendly solution for monitoring health and promoting physical activity.

## 3    Proposed System

The proposed work involves the integration of two modules: an Exercise-Activity Tracking Module and the Face Recognition Module. These modules collaborate to track walking activities within a community park by considering the specific constraints and parameters. The exercise activity tracking module focuses on monitoring and recording walking activities within the park, accounting for the defined perimeter and designated pathway for completing a lap. This module tracks individuals' movement along a pathway, capturing data related to their speed, distance covered, and duration of activity.

Simultaneously, the Face Recognition Module played a vital role in ensuring the identification and verification of individuals during walking sessions.

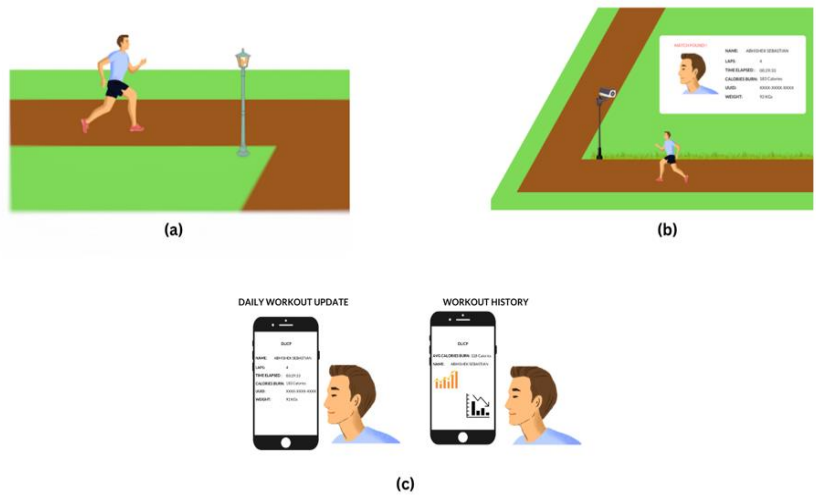

**Fig. 1.** Proposed workflow of DLICP

Fig. 1 visually depicts the workflow of DLICP, elucidating the sequential steps inherent in the process.

As Shown in Fig. 1(a), before commencing their workout, park users are only required to upload a photo of their face and provide basic details, such as their name and weight, to the DLICP system for seamless fitness tracking. As Shown in Fig. 1(b), DLICP utilizes advanced facial recognition and employs unique calorie calculation methods to enhance the accuracy of lap tracking and estimating calories burned, thereby providing users with a comprehensive and precise workout monitoring experience. In Fig. 1(c), DLICP allows users to effectively monitor their fitness progress, visualize workout data, and access stored information in the cloud, ensuring a convenient and informed fitness journey. It utilizes a facial recognition framework (as explained in subsection 3.1) to capture and analyze the faces of individuals participating in the activity. The module ensures that the subject's face is captured at least once during every lap around the park, allowing for the accurate monitoring and tracking of individual participation.

By combining the Exercise-Activity Tracking Module (as explained in Subsection 3.2) and the Face Recognition Module, a comprehensive system was established for tracking and monitoring walking activities within the community park. The integration of these modules ensures the synchronization of data on physical activity and individual identification, enabling a better analysis of individual performance, group statistics, and the overall impact of walking activities within the community park.

### 3.1 Face Recognition Module

In the proposed face recognition module, advanced techniques such as Multi-task Cascaded Convolutional Networks (MTCNN) [10], are employed for initial face detection and localization. Once a face is detected, the region of interest (ROI) is extracted by cropping the image. This ROI undergoes encoding or vectorization of facial features using a VGG-16 model, with its classifier head removed [11]. The VGG-16 model captures high-level features that represent the unique characteristics of the face.

To compare the generated face vectors with pre-existing encodings in the database, the cosine similarity metric [21] was used to measure the angular similarity between face vectors. Setting a suitable threshold above which cosine similarity is considered significant allows the verification of the face as belonging to a specific person.

In addition to MTCNN, a model trained specifically for faces, we explored multiple Generic Object Detection Models (YOLO), including YOLOv3, YOLOv5, YOLOv7, and YOLOv8. These models were trained with a diverse dataset of over 100,000 human face images, encompassing different racial backgrounds and lighting conditions. Our evaluation involved assessing their performance in terms of inference speed, average precision, and CPU memory usage.

The assessment (See Table 1) of inference speed and CPU memory usage was based on the single-core processing of the GPU, utilizing the NVIDIA Maxwell architecture with 128 NVIDIA CUDA® cores. In addition, the CPU features a quad-core ARM Cortex-A57 MP Core processor, and the memory is 4 GB 64-bit LPDDR4, operating at 1600MHz with a bandwidth of 25.6 GB/s.

**Table 1.** Comparative analysis of the different face detection models

| S.No | Model | Inference Speed (ms/face) | Average Precision (AP) | CPU Memory Usage (MB) |
|---|---|---|---|---|
| 1 | MTCNN | 50-100 | 0.88-0.92 | ~200 |
| 2 | YOLOv3 | 20-30 | 0.75-0.85 | ~500 |
| 3 | YOLOv5 | 10-20 | 0.80-0.90 | ~300 |
| 4 | YOLOv7 | 5-10 | 0.85-0.95 | ~250 |
| 5 | YOLOv8 | 3-5 | 0.90-0.97 | ~200 |

Our analysis revealed that a specific version, YOLOv8, outperformed the other versions by being the fastest with the least memory usage. This choice was necessitated by the need to address one of the constraints in real-time face detection and recognition for the DLICP.

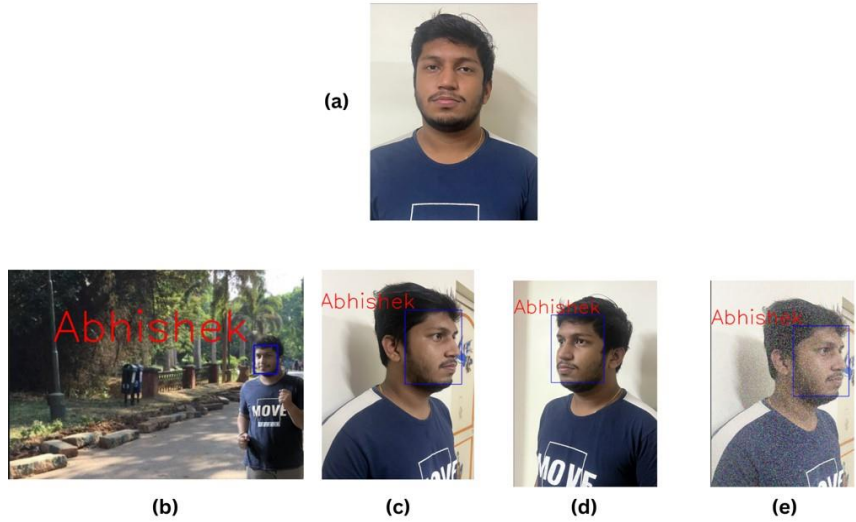

**Fig. 2.** Face recognition module analysis

Fig. 2(a) shows the input image to the framework, representing the original image used for the analysis. Fig. 2(b) shows the successful recognition of faces in a high-contrast environment, such as a park, highlighting the framework's effectiveness under challenging lighting conditions. Furthermore, Fig. 2(c) demonstrates the framework's capability to achieve successful recognition even when only the left profile of the subject's face is visible. Similarly, Fig. 2(d) shows successful recognition with only the right profile of the subject's face. These results emphasize the proficiency of the framework in identifying individuals from partial facial images. Finally, as shown in Fig. 2(e), the framework achieves successful recognition despite the presence of computer-generated noise simulating camera artifacts. This scenario demonstrates the robustness of the framework in handling image distortions and enhances its accuracy in face recognition tasks.

To conduct experiments with the subjects, we utilized the integrated YOLOv8, incorporating feature extraction through deep neural networks such as VGG-16. Accurate and reliable face verification was performed using a cosine similarity-based approach, which is of primary importance in the proper functioning of the DLICP.

### 3.2 Exercise-Activity Tracking Module

The Exercise-Activity Tracking module is designed to monitor and track physical exercise, specifically focusing on calculating the calories burned during the activity. This module uses a mathematical model to estimate calorie expenditure based on the average pace achieved during exercise. The average pace as shown in Eq. (2), is measured and updated each time the individual completes a lap in the park, which is detected through facial recognition by a camera.

$$Distance\ Covered = Predefined\ Park\ Perimeter\ X\ Number\ of\ laps \quad (1)$$

$$Average\ Pace = \frac{DistanceCovered}{TimeElapsed} \quad (2)$$

The calorie calculation process involves determining the Metabolic Equivalent of Task (MET) [12, 13] value based on the average pace achieved. According to the information in Table 2, MET values assigned to different walking activities depend on the attained average pace.

**Table 2.** MET for different types of walking activity w.r.t average pace

| S.No | Type of Walking Activity | Pace | MET |
|---|---|---|---|
| 1 | Strolling (Slow Walk) | < 5.6 Km / Hour | 2.0 |
| 2 | Brisk Walking | 5.6 – 6.4 Km / Hour | 5.0 |
| 3 | Concentrated Brisk Walking | 6.4 – 7.2 Km /Hour | 6.3 |
| 4 | Running | >11.2 Km / Hour | 11.5 |

These MET values serve as standardized indicators of energy expenditure associated with different walking activities, facilitating comparisons and assessments of physical exertion across individuals and studies [14].

The calories burned per minute [15] were calculated using Eq. (3). The total calories burned as shown in Eq. (4) during the exercise session were then obtained by multiplying the calories per minute by the elapsed time in minutes, considering the fractional part of an hour.

$$CaloriesBurnedPerMinute = \frac{MET\ X\ Subject\ Body\ Weight\ X3.5}{200} \quad (3)$$

$$Total\ Calories\ Burned = Calories\ Burned\ Per\ Minute\ X\ Time\ Elapsed \quad (4)$$

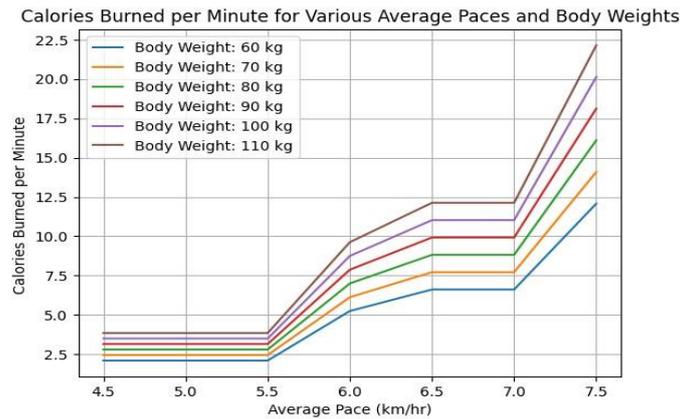

**Fig. 3.** Calories burned per minute for various average paces and body weights

The utilization of these MET values, derived from Table 1, allows for an accurate estimation of the calories burned during different walking activities based on the average pace achieved. Fig. 3 shows the relationship between average pace, body weights, and the corresponding calories burned per minute. This highlights how different combinations of average paces and body weights impact the rate of caloric expenditure during physical activities.

By considering these variations in MET values, the exercise-activity tracking module provides more precise and tailored calorie expenditure information for individuals engaged in various walking exercises.

## 4 Experiments and Results

### 4.1 Study Site Characteristics

In this section, we present the deployment of the proposed DLICP system at a community park in Porur, Chennai. The park features a closed one-path layout with a quadrilateral shape, offering a predefined perimeter of approximately 110 meters. To assess the effectiveness of the DLICP, we selected 22 subjects with diverse ages, body weights, and fitness levels (see Fig. 4).

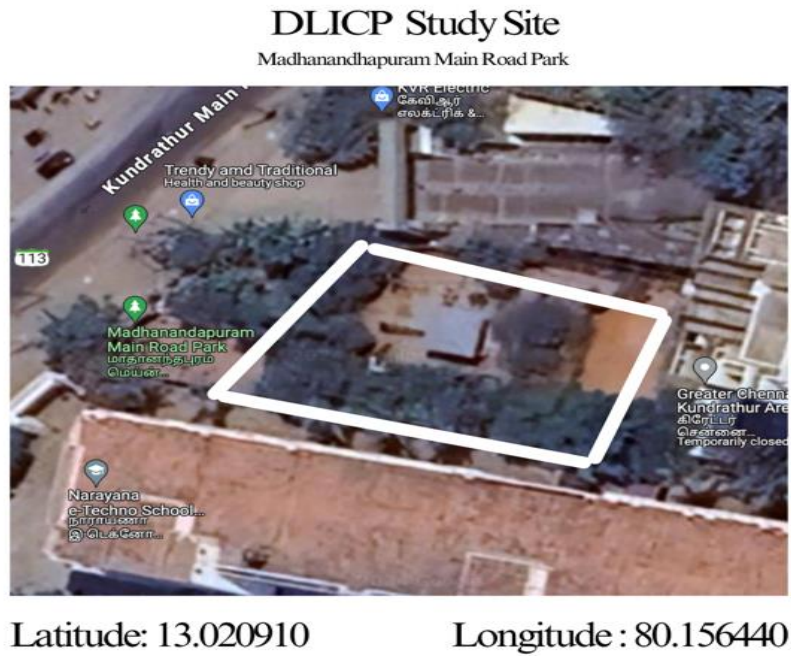

**Fig. 4.** Bird's eye view of the study site

## 4.2 Collection of Experimental Data

Our experiments aimed to evaluate performance of DLICP and compare it with the industry-standard Apple Watch Series 6. The first experiment focused on assessing the DLICP alone, with subjects engaging in their typical walking routine within the park. The DLICP walking activity measurement algorithm accurately calculated parameters such as calories burned, average pace, and distance covered by each subject. The collected data offered insights into individual workout attributes and served as a baseline for further analyses.

**Table 3.** Results of experiment 1 with DLICP

| Subject ID | Body Weight (in KGs) | Avg Pace (In Km/Hour) | MET | Calories Per Minute | Total Calories Burned |
|---|---|---|---|---|---|
| S1 | 73.3 | 6.3 | 5 | 6.41 | 192.41 |
| S2 | 74.2 | 6.1 | 5 | 6.49 | 194.77 |
| S3 | 66.7 | 7 | 6.3 | 7.35 | 220.61 |
| S4 | 88 | 5 | 2 | 3.08 | 92.40 |
| S5 | 78.6 | 6.1 | 5 | 6.87 | 206.32 |
| S6 | 78.2 | 6.28 | 5 | 6.84 | 205.27 |
| S7 | 85.7 | 5.1 | 2 | 2.99 | 89.98 |
| S8 | 78.4 | 4.8 | 2 | 2.74 | 82.32 |
| S9 | 49.5 | 8.8 | 11.5 | 9.96 | 298.85 |
| S10 | 81.8 | 5.8 | 5 | 7.15 | 214.72 |
| S11 | 77.4 | 7.1 | 6.3 | 8.53 | 256.00 |
| S12 | 66.9 | 5.9 | 5 | 5.85 | 175.61 |
| S13 | 73.9 | 5.1 | 2 | 2.58 | 77.59 |
| S14 | 74.8 | 5.3 | 2 | 2.61 | 78.54 |
| S15 | 97.3 | 7.1 | 6.3 | 10.72 | 321.81 |
| S16 | 81 | 6.2 | 5 | 7.08 | 212.62 |
| S17 | 98 | 5.3 | 2 | 3.43 | 102.90 |
| S18 | 75.7 | 4.9 | 2 | 2.64 | 79.48 |
| S19 | 70.5 | 6 | 5 | 6.16 | 185.06 |
| S20 | 78.1 | 3.8 | 2 | 2.73 | 82.00 |
| S21 | 86.2 | 4.2 | 2 | 3.01 | 90.51 |
| S22 | 77.8 | 5.4 | 2 | 2.72 | 81.69 |

In the second experiment, we conducted a comparative analysis between DLICP and Apple Watch Series 6. For this purpose, we set the Apple Watch to the "Walking Activity" mode within the fitness app. Both the DLICP and Apple Watch measured the calories burned during the subjects' walking routines. By comparing the recorded values, we evaluated the relative accuracy of the DLICP compared to the industry-standard Apple Watch.

**Table 4.** Comparative analysis of DLICP and apple watch

| Subject ID | DLICP (In Calories) | Apple Watch (In Calories) | Deviation (In Calories) |
|---|---|---|---|
| S1 | 192.41 | 203 | -10.59 |
| S2 | 194.77 | 186 | +8.77 |
| S3 | 220.61 | 215 | +5.61 |
| S4 | 92.40 | 87 | +5.4 |
| S5 | 206.32 | 200 | +6.32 |
| S6 | 205.27 | 198 | +7.27 |
| S7 | 89.98 | 95 | -5.02 |
| S8 | 82.32 | 86 | -3.68 |
| S9 | 298.85 | 295 | +3.85 |
| S10 | 214.72 | 209 | +5.72 |
| S11 | 256.00 | 263 | -7 |
| S12 | 175.61 | 177 | -1.39 |
| S13 | 77.59 | 79 | -1.41 |
| S14 | 78.54 | 75 | 3.54 |
| S15 | 321.81 | 300 | +21.81 |
| S16 | 212.62 | 207 | +5.62 |
| S17 | 102.90 | 100 | +2.9 |
| S18 | 79.48 | 73 | +6.48 |
| S19 | 185.06 | 177 | +8.06 |
| S20 | 82.00 | 84 | -2.00 |
| S21 | 90.51 | 86 | +4.51 |
| S22 | 81.69 | 77 | +4.69 |

In Experiments 1 and 2, the duration was limited to 30 min. This time constraint was imposed to ensure consistency and to effectively manage the experimental process. By capping the elapsed time, we aimed to capture a representative snapshot of subjects' walking routines within a reasonable timeframe.

### 4.3 Interpreting Experimental Data

The results of the experiments (see Table 3 and 4) demonstrate the effectiveness of DLICP in measuring the workout attributes. The DLICP algorithm successfully calculated the calories burned, average pace, and distance covered for the subjects, considering their unique attributes and fitness levels.

$$MAE = \frac{1}{N}\sum_{i=1}^{N}|DLICP_i - Apple\ Watch_i| \quad (5)$$

$$MPE = \frac{1}{N}\sum_{i=1}^{N}\left(\frac{DLICP_i - Apple\ Watch_i}{DLICP_i}\right) X\ 100 \quad (6)$$

The performance of DLICP in measuring workout attributes, compared to the Apple Watch, can be evaluated quantitatively using metrics such as the Mean Absolute Error (MAE) as shown in Eq. (5) and Mean Percentage Error (MPE) as shown in Eq. (6). The MAE value of 5.64 indicates an average absolute difference of 5.64 calories between the DLICP and Apple Watch measurements. Similarly, the MPE value of 1.96% signifies that, on average, the DLICP measurements deviate by 1.96% from the Apple Watch measurements in terms of calorie estimation. These results demonstrate that DLICP provides a reasonably accurate estimation of workout attributes with relatively low deviations from the measurements obtained by the Apple Watch.

In the evaluation of DLICP performance, certain subjects displayed a commendable alignment between the estimated and observed calorie expenditure. Notably, individuals S3, S9, S11, and S15 exhibited relatively high Total Calories Burned, suggesting the effectiveness of DLICP in capturing their diverse activity levels. However, challenges arose when assessing subject S22, as DLICP yielded notably lower Total Calories Burned, suggesting potential limitations in accurately estimating calorie expenditure for the specific individual. This discrepancy was attributed to his lower average pace compared to that of the other subjects. This pattern extended to S18, S21, S4, and S14.

Upon comparison with readings from the smartwatch, it was observed that the readings were similar or closely aligned with minimal deviations. Hence, when the average pace dramatically decreased, both the smartwatch and DLICP reflected a reduction in the calories burned at end of the activity.

Another intriguing observation pertains to the influence of Metabolic Equivalent of Task (MET) values in DLICP estimations, as evidenced by subject S9, possessing MET values of 11.5, resulting in notably higher Total Calories Burned. This is because, from observation, Subject S9 engaged in high-paced running for the majority of the stipulated experimental time limit. This insight might help us understand the underlying truth: the more a subject engages in high-intensity running during the exercise, the more calories they will burn eventually at the end of the event.

Pronounced divergences emerged in S7, S8, and S20. In these cases, discrepancies of -5.02, -3.68, and -2.00, respectively, suggest potential intricacies or constraints in DLICP measurements. It is conceivable that factors dependent on the subjects' heart rate measurements and Apple Watch's proprietary algorithm could have contributed to these variations.

Factors such as the individual's body weight and average pace play a pivotal role in calorie estimation, contributing to the differences between DLICP and the Apple Watch. The inherent variability in these parameters underscores the complexity of accurately gauging calorie expenditure during PA.

Moreover, the distance calculated in the watch, which relies on accelerometer readings and GPS, may not be entirely accurate. This can influence the proprietary algorithms in the watch, affecting the final calorie burned readings.

It's crucial to acknowledge that a universally acknowledged standard or state-of-the-art (SOTA) method to definitively determine Calorie Burn is absent. The absence of a standardized benchmark emphasizes the need for continuous research and refinement to develop robust algorithms that can effectively account for the multifaceted nature of individual physiological responses to varying activities.

While DLICP demonstrates an overall alignment with Apple Watch estimates for many subjects, the presence of both overestimation and underestimation underscores the need for further investigation into the specific characteristics of subjects and activities contributing to these discrepancies.

Considering the tradeoffs between the two systems, the absence of heart rate measurements in DLICP may result in less accuracy. Relying on distance approximation using accelerometers and GPS in the Apple Watch might introduce uncertainties in determining the precise average pace, thereby impacting the overall accuracy.

Nevertheless, exploring non-invasive techniques, such as microwave radars for heart BPM measurement, along with the existing DLICP architecture, presents an opportunity to enhance accuracy. By incorporating newer proportionality constants, we can refine the existing mathematical model and improve the overall precision of the calorie expenditure estimation. This suggests that DLICP might represent an initial step towards providing reasonable results for non-invasive calorie expenditure estimations. However, the call for refined algorithms is paramount for enhancing the accuracy of these estimations. Continued research with diverse subjects and nuanced considerations for further algorithm development holds promise for advancing the precision and reliability of such estimation methods in future research.

## 5    Conclusion

In conclusion, this study underscores the efficacy and precision of the DLICP algorithm for measuring workout attributes with notable accuracy. Through meticulous experimentation and comprehensive comparison with Apple Watch Series 6, DLICP has demonstrated its ability to accurately calculate calories burned, average pace, and distance covered. This precision extends across individuals with diverse attributes and varying fitness levels, as evidenced by the recorded data, with minimal deviations from the industry standard. The mean absolute error (MAE) of 5.64 calories and mean percentage error (MPE) of 1.96% affirm the reliability of the DLICP, positioning it as a valuable tool for assessing workout performance. Future research should explore the application of DLICP in larger and more diverse populations to further validate its accuracy and reliability in various real-world scenarios.

## Ethical Approval

All activities conducted in studies involving human participants were conducted in accordance with the ethical standards set by the institutional and/or national research committee. The procedures aligned with the principles of the 1964 Declaration of Helsinki and its subsequent amendments, or equivalent ethical standards.

## Conflicts of Interests

No conflicts of interest were reported.

## Credit Author Statement

Abhishek Sebastian - Conceptualization, Methodology, Software, Writing – Original Draft, Project administration.
    Annis Fathima A - Supervision, Writing – Review & Editing.
    Pragna R - Conceptualization, Methodology, Writing – Original Draft.
    Madhan Kumar S - Formal analysis, Investigation, Software and Data Curation.
    Jesher Joshua M – Software, Visualization and Formal analysis.

## Code Availability

Yes, Custom Code.

## Data Availability

Data Available on Request.

## Funding Statement

No funding from any institution was utilized for this research.

## Acknowledgment

The researchers express their gratitude to the local park authorities and corresponding officials for their support during the tenure of the study.